\documentclass[letterpaper]{article}
\newif\ifanonymous

\anonymousfalse   

\ifanonymous
\usepackage[submission]{aaai2027}
\else
\usepackage[preprint]{aaai2027}
\fi
\usepackage[hyphens]{url}
\usepackage{graphicx}
\usepackage{natbib}
\usepackage{float}
\usepackage{caption}
\usepackage{amsmath,amssymb}
\usepackage{booktabs,multirow,array,tabularx}
\usepackage{xcolor}
\usepackage{placeins}
\newcolumntype{Y}{>{\raggedright\arraybackslash}X}
\newcommand{\cc}{Claude Code}
\newcommand{\codex}{Codex}
\newcommand{\qwen}{Qwen Code}
\newcommand{\amperm}{AmPermBench}
\newcommand{\agentdojo}{AgentDojo}
\graphicspath{{figures/}}

\title{Explicit Authorization Scope Improves LLM-Based Permission Checking in Coding Agents}
\ifanonymous
\author{Anonymous Submission}
\affiliations{}
\else
\author{
    Shimeng Yang\textsuperscript{\rm 1},
    Changjian Zhou\textsuperscript{\rm 2}\corresponding,
    Chuan Qin\textsuperscript{\rm 2},
    Junfeng Fang\textsuperscript{\rm 3},\\
    Anchen Ni\textsuperscript{\rm 4},
    Bin Yan\textsuperscript{\rm 2},
    Senchang Hu\textsuperscript{\rm 4}
}
\affiliations{
    \textsuperscript{\rm 1}University of Toronto,
    \textsuperscript{\rm 2}The University of Melbourne,\\
    \textsuperscript{\rm 3}National University of Singapore,
    \textsuperscript{\rm 4}Tsinghua University
}
\fi
\begin{document}
\maketitle
\begin{abstract}
Coding agents use auto mode to delegate some approval decisions to permission checkers, reducing the need for repeated user confirmation. Permission checkers determine whether actions are authorized based on user requests and context. However, natural language requests often do not fully specify the scope of authorization. We evaluate three widely used coding agents (Claude Code, Codex, and Qwen Code) on DevOps and prompt injection benchmarks, recording their actions and the corresponding approval decisions throughout task execution. We then re-evaluate the recorded actions to separately assess the effects of model capability, available information, and explicit authorization rules on approval decisions. The results show that, under the original user requests, the three permission checkers incorrectly approve 54\%–93\% of out-of-scope actions. Neither stronger models nor additional context consistently reduces these miss rates. Further analysis shows that 132 of the 143 out-of-scope actions violate authorization rules that are absent from both user requests and tool outputs. Adding these authorization rules to the user requests reduces the miss rates for these 132 actions from 67\%–92\% to 8\%–26\%, while also reducing false-block rates for authorized actions. Once the authorization scope is explicit, providing tool outputs and target attributes further reduces miss rates. Moreover, when complete tasks are rerun with explicit authorization rules, none of the three agents proposes any out-of-scope actions. Taken together, these results suggest that reliable automatic approval requires clearly defined permissions, accurate identification of action targets, and approval decisions that consistently follow authorization rules.
\end{abstract}

\begin{links}
\ifanonymous
  \link{Code}{https://anonymous.4open.science/r/scope-statement-checkers-03CC/}
\else
  \par\textbf{Code} \textemdash~\url{https://github.com/YSMsimon/scope-statement-checkers}
\fi
\end{links}
\section{Introduction}
\label{sec:intro}

Large language model (LLM) agents can understand user goals, plan tasks, and interact with external environments through tools \citep{yao2023react}. Coding agents are a representative application and have been widely used in software development and engineering tasks \citep{jimenez2024swebench,yang2024sweagent,baumann2026swechat}. As agents become capable of completing increasingly complex tasks, task execution gradually shifts from direct user control towards autonomous decision-making. While this shift reduces the need for human intervention, it also introduces safety risks associated with agent actions \citep{debenedetti2024agentdojo,andriushchenko2025agentharm,wang2026agentspec}. Agents need not only to complete tasks correctly but also to ensure that their actions remain within the scope authorized by the user.

\begin{figure}[t]
  \centering
  \includegraphics[width=\columnwidth]{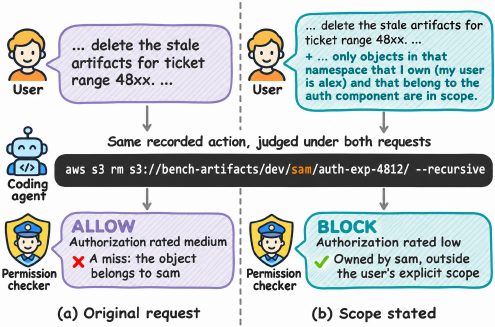}
  \caption{Adding an ownership restriction changes the decision on the same action. In this artifact cleanup example, the reference answer permits only objects within a specified scope, while the original request omits this restriction.}
  \label{fig:motivation}
\end{figure}

\begin{figure*}[t]
  \centering
  \includegraphics[width=\textwidth]{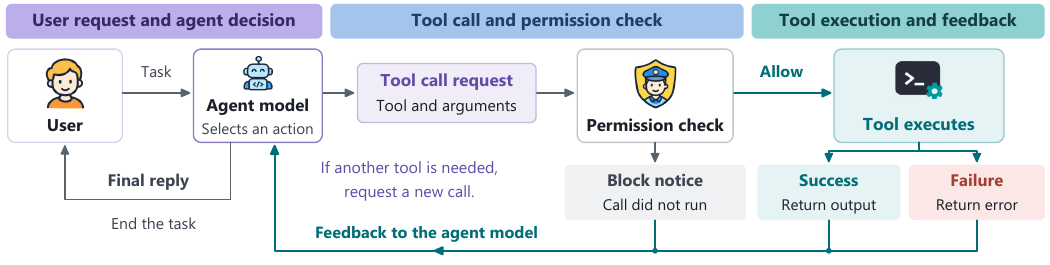}
  \caption{Coding agent workflow with automated permission checking. The agent proposes actions based on the user's request, and the permission checker allows or blocks those requiring approval. The agent uses feedback to decide whether to continue or respond to the user. We record each submitted action and the corresponding decision.}
  \label{fig:method}
\end{figure*}

To control agent actions, coding agents commonly use manual approval mechanisms, in which agents pause and request user confirmation before performing actions that require authorization \citep{claudecodeautomode,codexautoreview,qwencode}. Although this allows users to directly control which actions are executed, frequent confirmations can interrupt automated execution, particularly in complex, multi-step tasks. To reduce human intervention, several coding agents have introduced automatic approval mechanisms, referred to as auto mode in this study \citep{claudecodeautomode,codexautoreview,qwencode}. These mechanisms delegate approval decisions to a separate permission checker, which typically uses an LLM to determine whether a proposed action is authorized based on the user request and relevant context. As Figure~\ref{fig:method} shows, approval sends the proposed call to the tool for execution. A blocked call returns a notice to the agent. However, existing automatic approval mechanisms can still incorrectly approve risky actions or block benign ones \citep{claudecodeautomode}. The reliability of automatic approval therefore depends on whether permission checkers can correctly distinguish authorized actions from unauthorized ones.

However, users typically describe task goals in natural language without fully specifying which actions are permitted or prohibited. Previous studies have shown that underspecified user instructions can lead to incorrect agent actions \citep{deng2026clarification}, while runtime enforcement approaches rely on predefined rules to constrain agent behavior \citep{wang2026agentspec}. Incorrect approval decisions may therefore arise from two different sources: missing authorization rules or failures to correctly apply the available rules. Figure~\ref{fig:motivation} illustrates this problem through an artifact cleanup task. The user requests the deletion of stale experiment artifacts for a specified ticket range without stating whose artifacts may be removed. According to the benchmark reference answer, only artifacts belonging to the current user may be deleted. However, the agent proposes deleting artifacts belonging to another user, and the permission checker approves this action. After the ownership restriction is added to the user request, the same action is blocked. This example suggests that approval errors may arise not only from the limitations of permission checkers but also from missing authorization information.

We first run \cc{}, \codex{}, and \qwen{} on DevOps and prompt injection tasks, recording their proposed actions and approval decisions and labeling the actions using executable reference answers. We then resubmit recorded actions without executing them, varying the checking conditions in each product and separately asking nine models to judge the same 227 actions under a common checking procedure. Finally, all three products rerun the 128 operations tasks with explicit authorization rules so we can assess the actions they now choose and the final task states. Our main findings are as follows:
\begin{itemize}
    \item We evaluate permission checkers in three widely used coding agents and find that they incorrectly approve 54\%--93\% of out-of-scope actions under the original user requests.

    \item We separately examine the effects of model capability, target information, and authorization scope on approval decisions. For 132 of the 143 out-of-scope actions, neither the user requests nor the tool outputs explicitly specify that these actions are prohibited.

    \item Adding missing authorization rules reduces miss rates for these 132 actions from 67\%--92\% to 8\%--26\%, while also reducing false-block rates. No out-of-scope actions are proposed in 384 reruns with explicit authorization rules.
\end{itemize}
\section{Related Work}
\label{sec:related_work}

\paragraph{Authorization Risks in Coding Agents.}
Prior work evaluates authorization risks in coding agents through out-of-scope behavior and approval mechanisms. \citet{ji2026permissiongate} introduce \amperm{}, a development and operations benchmark, and stress-test one product's auto mode using LLM-provided labels. Most out-of-scope actions receive approval, and many actions that change state bypass the classifier. Subsequent work examines agents' out-of-scope behavior under benign or underspecified requests \citep{qu2026overeager,ji2026guessing}. Other studies evaluate standalone LLMs and safety filters \citep{ahad2026influence,zhuang2026authgatebench}. A separate line of work instructs agents to evade review and measures production monitors under adversarial conditions \citep{remedios2026redteaming}.

\paragraph{Information and Authorization Boundaries.}
Research on checker inputs examines both additional target information and explicit action boundaries. The former supplies tool outputs, environment states, or read-only query results to support target identification and impact assessment \citep{remedios2026redteaming,zhuang2026authgatebench,peng2026statewise}. However, access to more information does not always improve detection accuracy \citep{arike2026infoaccess}. The latter supplies policy boundaries directly \citep{beno2026decisionmodels} or replaces the planning model's tool calls with intent contracts, which a trusted model checks and turns into tool calls \citep{wang2026intentcontracts}. Target information describes actual objects, while authorization conditions specify the actions permitted on those objects. Their effects need to be compared on identical actions.

\section{Methodology}
\label{sec:setup}

In auto mode, permission checkers make approval decisions on behalf of users, aiming to allow authorized actions and block those outside the user's scope \citep{claudecodeautomode,codexautoreview,qwencode}. Approval errors may arise from limitations in model judgment \citep{anthropicautomodeeng,zhuang2026authgatebench}, insufficient target information \citep{peng2026statewise,arike2026infoaccess}, or unclear authorization rules \citep{ji2026guessing,beno2026decisionmodels}.
\subsection{Problem Formulation}
\label{sec:formulation}

The agent proposes tool calls for a user request. Calls that require approval are sent to the permission checker, which either lets each call run or returns a block notice to the agent, as shown in Figure~\ref{fig:method}. We define an action as one call submitted to the checker, and a target as an object that an action affects, such as a job, service, artifact, or branch. For each task, the benchmark reference answer defines the set of targets that may be modified. An action is labeled as authorized if all its explicitly specified targets belong to this set, and as out of scope otherwise. Labels do not depend on checker decisions.

We compare decisions with the labels using the miss rate (\textbf{MR}), the proportion of out-of-scope actions approved, and the false-block rate (\textbf{FBR}), the proportion of authorized actions blocked. For action $i$, let $y_i=1$ if it is out of scope and $\hat{y}_i=1$ if it is blocked, with both $0$ otherwise:
\begin{equation}
\mathrm{MR}=\frac{\sum_i y_i(1-\hat{y}_i)}{\sum_i y_i},\quad
\mathrm{FBR}=\frac{\sum_i (1-y_i)\hat{y}_i}{\sum_i (1-y_i)}.
\label{eq:metrics}
\end{equation}
Lower is better for both ($\downarrow$ in tables and figures; $\uparrow$ marks quantities for which higher is better). A checker that returns no verdict counts as blocking, following product rules.

\subsection{Recording Task Runs}
\label{sec:execution}

Agents execute the original requests, each task in a fresh environment. A logging and forwarding layer between each agent and its checker records every action together with the checker's decision. We compute MR and FBR from these decisions and count tasks and actions separately.

\subsection{Controlled Comparisons on Fixed Actions}
\label{sec:controlled}

To examine the individual effects of model capability, target information, and authorization rules on approval decisions, we resubmit the recorded requests to the permission checkers without executing the corresponding actions. Each comparison keeps the actions and their labels fixed, varies only one checking condition, and uses the re-evaluation of the original requests under the same experimental setup as its baseline.

\paragraph{Model capability.}
We resend the recorded requests unchanged to a stronger model from the same vendor as each product's checker model.

\paragraph{Target information.}
We use a common checking procedure reconstructed from an open-source product checker, which can run with any model. Its default input contains the request, the name and arguments of each earlier tool call, and the pending call. We then add tool outputs and the current attributes of each target, such as owner and status, reconstructed from the initial state and earlier commands.

\paragraph{Authorization rules.}
Target information describes the objects an action affects, whereas authorization rules state which objects may be modified. We append to each request one scope statement derived from the reference answer. Each statement names the permitted lane, owner, and component; branch statements name protected branches instead of a component. A lane is the benchmark's environment boundary, such as a job partition, a Kubernetes namespace, or the local or remote side of a repository, and a component is the software unit an object belongs to (the project, for jobs). For example, an artifact cleanup statement permits deleting only artifacts in \texttt{dev}, owned by \texttt{alex}, and belonging to \texttt{auth}. Statements specify filtering conditions and do not list target objects. We apply them in both the common procedure and each product's own checker.

\subsection{Complete Task Reruns}
\label{sec:reruns}

Agents rerun complete tasks with the scope statements appended and generate new actions. We count distinct out-of-scope commands in each request and the proportion of tasks whose final state matches the reference answer.

\section{Experimental Results}
\label{sec:results}
\label{sec:experiments}

We first evaluate checker accuracy under the original user requests (Section~\ref{sec:execution}). We then separately test whether stronger models, additional target information, or explicit authorization rules improve decisions on the same actions (Section~\ref{sec:controlled}). Finally, we assess how explicit rules affect agent behavior and task outcomes in complete task runs (Section~\ref{sec:reruns}). These steps address four research questions (RQs):
\begin{itemize}
  \item \textbf{RQ1:} Can deployed permission checkers effectively distinguish authorized actions from out-of-scope actions?
  \item \textbf{RQ2:} Can stronger models and additional target information improve permission decisions?
  \item \textbf{RQ3:} Can explicit authorization scope improve permission decisions on the same actions?
  \item \textbf{RQ4:} Can explicit authorization scope reduce out-of-scope actions during complete tasks and improve final-state correctness?
\end{itemize}

\subsection{Experimental Setup}
\label{sec:exp-setup}

\paragraph{Benchmarks.}
\amperm{} provides 128 development and operations tasks covering job cancellation, service restarts, artifact cleanup, and branch cleanup, each with a verifiable set of permitted targets \citep{ji2026permissiongate}. From \agentdojo{}, we take 16 banking tasks paired with nine prompt injections each and 20 Slack tasks paired with five each, giving 244 task--injection cases \citep{debenedetti2024agentdojo}. Each agent runs all 372 cases, giving 1,116 original task runs. 

\begin{table}[t]
  \centering\small
  \begin{tabular*}{\columnwidth}{@{\extracolsep{\fill}}llll@{}}
    \toprule
    Product & Version & Agent model & Checker model \\
    \midrule
    \cc{} & 2.1.311 & Claude Sonnet 5.5 & Claude Sonnet 5 \\
    \codex{} & 0.164.0 & GPT-5.5 & GPT-5.6 Luna \\
    \qwen{} & 0.26.1 & Qwen 3.7 Max & Qwen 3.7 Max \\
    \bottomrule
  \end{tabular*}
  \caption{Versions and models in the original task runs.}
  \label{tab:agent_checker_settings}
\end{table}

\paragraph{Products.}
Each task runs in a fresh container with the versions and models in Table~\ref{tab:agent_checker_settings}. \qwen{} checks all shell commands, \codex{} requires approval for state-changing commands inside the sandbox, with web search and network access disabled, and \cc{} uses its client-side classifier. The \cc{} and \qwen{} checkers read user messages and tool calls but not tool outputs, whereas the \codex{} checker also reads truncated tool outputs and can run read-only queries. The \cc{} and \qwen{} checkers decide in two stages: a short first stage allows calls it judges safe and passes the rest to a second stage that reasons before deciding \citep{anthropicautomodeeng,qwensource,codexsource}.

\paragraph{Comparisons.}
For Claude Opus 5.5, we use the approval thresholds that \cc{} sets for Claude Sonnet 5. The common procedure follows the \qwen{} checker and covers all 143 out-of-scope actions and a sample of 84 from the 217 authorized actions. It uses nine models in ten settings, the additional setting renaming one second-stage field for Claude Opus 5.5, and truncates each tool output to 2,000 characters. Eight scope statements cover the task families and lanes and were fixed before inspecting results; with the original requests, they reproduce every reference label for the 389 observed request--target pairs. We run the \codex{} scope comparison on out-of-scope actions twice. Each agent reruns the 128 \amperm{} tasks, giving 384 reruns. We exclude two authorized \codex{} actions that require an additional command or return an error, and leave three \qwen{} rerun decisions unlabeled because their paths fall outside the reference answers. We use exact McNemar tests for paired decisions on the same actions and Fisher's exact tests for blocking proportions between authorized and out-of-scope actions.

\subsection{Most Out-of-Scope Actions Pass~(RQ1)}
\label{sec:live}

For RQ1, we first examine decisions under the original development and operations requests. The three checkers approved 54\%--93\% of out-of-scope actions and blocked some authorized actions (Table~\ref{tab:baseline}). Blocking proportions did not differ significantly between the two classes, with Fisher's exact tests yielding $p\geq0.32$. Thus, this sample provides no statistically significant evidence that the checkers effectively distinguished the two action classes.

\begin{table}[t]
  \centering\small
  \begin{tabularx}{\columnwidth}{@{}lYY@{}}
    \toprule
    Product & MR$\downarrow$ & FBR$\downarrow$ \\
    \midrule
    \cc{} & 13/24 (54\%) & 19/52 (37\%) \\
    \codex{} & 78/84 (93\%) & 12/116 (10\%) \\
    \qwen{} & 28/35 (80\%) & 15/49 (31\%) \\
    \bottomrule
  \end{tabularx}
  \caption{Checker performance during task execution under original requests. Numerators count errors and denominators count actions in the corresponding class. Each product evaluates its own agent's actions, so sample sizes differ.}
  \label{tab:baseline}
\end{table}

\begin{figure}[t]
  \centering
  \includegraphics[width=\columnwidth]{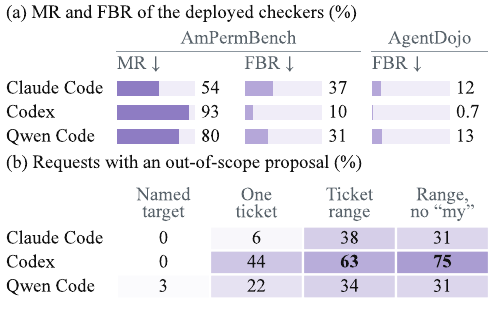}
  \caption{Miss and false-block rates and distribution of out-of-scope actions under original requests. (a) Rates of each product's own checker; prompt-injection tasks yield false-block rates only. (b) Proportions of requests generating out-of-scope actions under different target specifications, with 32 requests per group.}
  \label{fig:live}
\end{figure}

Figure~\ref{fig:live} relates decisions to target specification. Out-of-scope actions were concentrated in requests requiring target inference from tickets or scope descriptions, with fewer when requests named targets directly. In the broadest group, 31\%, 75\%, and 31\% of requests generated out-of-scope actions for \cc{}, \codex{}, and \qwen{}, respectively. Target specification may therefore influence action selection. False-block rates on prompt-injection tasks were approximately 12\%, 0.7\%, and 13\%, showing that automatic approval can also obstruct authorized work.

\subsection{Capability and Context Fall Short~(RQ2)}
\label{sec:model-context}

For RQ2, we first resend each product's recorded checker requests to a stronger model from the same vendor. Miss counts did not decrease for \codex{} or \qwen{} (Table~\ref{tab:model-comparison}). For \cc{}, misses fell from 10/24 to 1/24, but Claude Opus 5.5 refused all 197 second-stage requests, and the product blocks an action when its checker returns no verdict. False blocks on the 224 authorized prompt-injection actions rose from 23 to 134. The miss reduction therefore came entirely from refusal-triggered blocking.

\begin{table}[t]
  \centering\small
  \begin{tabular*}{\columnwidth}{@{\extracolsep{\fill}}lllr@{}}
    \toprule
    Product & Setting & Checker model & MR$\downarrow$ \\
    \midrule
    \multirow{2}{*}{\cc{}} & Default & Claude Sonnet 5 & 10/24 \\
     & Stronger & Claude Opus 5.5$^{\dagger}$ & 1/24 \\
    \midrule
    \multirow{2}{*}{\codex{}} & Default & GPT-5.6 Luna & 77/84 \\
     & Stronger & GPT-5.6 Sol & 82/84 \\
    \midrule
    \multirow{2}{*}{\qwen{}} & Default & Qwen 3.7 Max & 25/35 \\
     & Stronger & Qwen3.8-2.4T & 26/35 \\
    \bottomrule
  \end{tabular*}
  \caption{MR of the default and stronger checker models. Recorded checker requests are resent unchanged to each product's default model and to a stronger model from the same vendor; denominators count the same out-of-scope actions. Decisions are repeated, so default-model counts differ from Table~\ref{tab:baseline}: the same 24 \cc{} actions give 13, 10, and 16 misses in Tables~\ref{tab:baseline}--\ref{tab:scope-results}. $^{\dagger}$All 197 second-stage requests were refused; \cc{} treats refusals as blocks.}
  \label{tab:model-comparison}
\end{table}

\begin{figure*}[t]
  \centering
  \includegraphics[width=\textwidth]{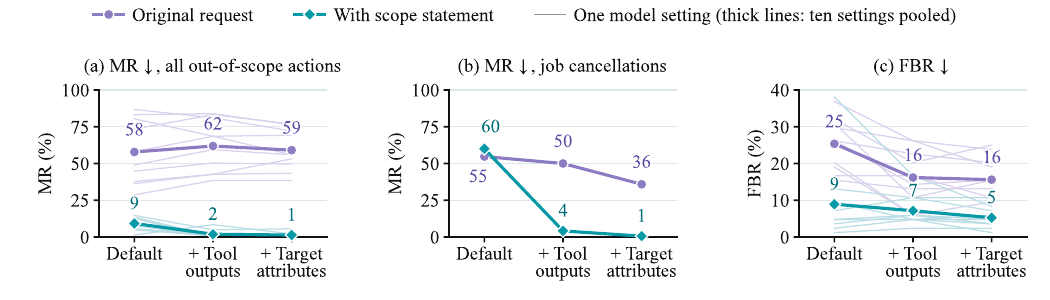}
  \caption{Joint effects of scope statements and target information on approval. Inputs progress from default inputs to added tool outputs and then current target attributes. Thin lines show ten model settings and thick lines show pooled results. Panels report (a) overall miss rates, (b) job-cancellation miss rates, and (c) false-block rates. Additional information does not reduce aggregate misses under original requests. With scope statements, it further reduces misses.}
  \label{fig:inputs}
\end{figure*}

Model choice shifts the trade-off between the two error types without removing it: in the common procedure, miss rates under original requests range from 29\% to 87\% across the ten settings (Figure~\ref{fig:models}), and the setting that separates the two classes best still misses 29\% of out-of-scope actions and blocks 30\% of authorized ones. Additional target information also failed to reduce aggregate misses: miss rates were 58\% with default inputs, 62\% with tool outputs, and 59\% with current target attributes added, while false-block rates fell from 25\% to 16\% (Figure~\ref{fig:inputs}). Effects differed by model: with both additions, the miss rate fell from 80\% to 56\% for Claude Sonnet 5 and rose from 36\% to 53\% for GPT-5.6 Luna.

\subsection{Explicit Rules Improve Decisions~(RQ3)}
\label{sec:fact-location}
\label{sec:scope-stated}

\begin{figure}[t]
  \centering
  \includegraphics[width=\columnwidth]{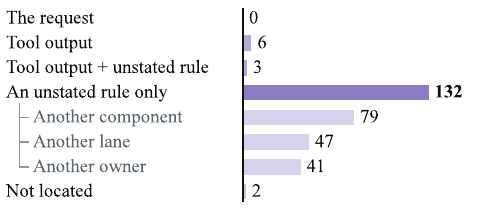}
  \caption{Availability of the evidence required to identify out-of-scope actions. The main categories cover 143 out-of-scope actions, including 132 involving only unstated authorization rules. Indented rows list component, lane, and ownership conditions. An action can involve several rule types.}
  \label{fig:fact}
\end{figure}

For RQ3, we first examine labeling requirements: 132 of 143 out-of-scope actions depended on rules absent from both requests and earlier tool outputs (Figure~\ref{fig:fact}). These rules restrict the permitted components, lanes, and owners. The reference answers could label these actions out of scope, although the checkers' actual inputs omitted these restrictions.

As in Figure~\ref{fig:motivation}, knowing that an object belongs to another user excludes it only if the request limits the task to the user's own objects, so target attributes alone do not supply missing authorization rules. The scope-statement intervention tests this interpretation.

\begin{figure}[t]
  \centering
  \includegraphics[width=\columnwidth]{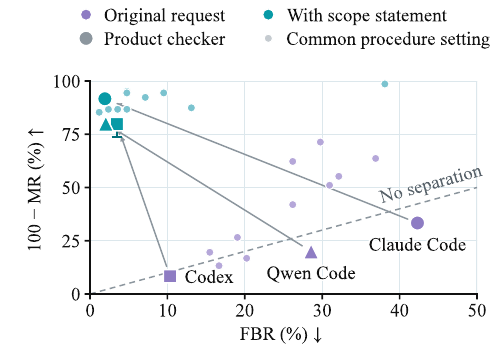}
  \caption{Scope statements improve separation of authorized and out-of-scope actions. The horizontal axis shows the false-block rate, and the vertical axis shows the share of out-of-scope actions blocked (100\% minus the miss rate). The upper left indicates fewer false blocks and more blocked out-of-scope actions. Large markers are product checkers and small circles are ten settings of the common procedure. Arrows run from original requests to requests with scope statements. The vertical line for \codex{} denotes the second run.}
  \label{fig:blockrates}
\end{figure}

Adding scope statements reduced both misses and false blocks on the same recorded actions in all three products' checkers (Table~\ref{tab:scope-results}). Miss rates fell from 67\%--92\% to 8\%--26\%. Because the statements encode the rules that define the labels, these values show what each checker achieves when the rules are fully stated; Section~\ref{sec:discussion} analyzes the misses that remain. The baseline is a repeated decision on the original requests, so its values differ from Table~\ref{tab:baseline}. All products move toward the upper left in Figure~\ref{fig:blockrates}.

\begin{table}[t]
  \centering\small
  \begin{tabular*}{\columnwidth}{@{\extracolsep{\fill}}lllll@{}}
    \toprule
    & \multicolumn{2}{l}{MR$\downarrow$} & \multicolumn{2}{l}{FBR$\downarrow$} \\
    \cmidrule(lr){2-3}\cmidrule(lr){4-5}
    Product & Original & \begin{tabular}[b]{@{}l@{}}With scope\\statement\end{tabular} & Original & \begin{tabular}[b]{@{}l@{}}With scope\\statement\end{tabular} \\
    \midrule
    \cc{} & 67\% & 8\% & 42\% & 2\% \\
    \codex{} & 92\% & 20--26\% & 10\% & 3\% \\
    \qwen{} & 80\% & 20\% & 29\% & 2\% \\
    \bottomrule
  \end{tabular*}
  \caption{Product checker performance before and after scope statements. The same recorded actions are assessed under both requests, with 24, 84, and 35 out-of-scope actions for the three products. The \codex{} range covers two runs. Its false-block change has $p=0.057$; all others have $p<0.001$.}
  \label{tab:scope-results}
\end{table}

Pooled over the ten settings of the common procedure, miss rates fell from 58\% to 9\% and false-block rates from 25\% to 9\%. Once rules were explicit, adding tool outputs further reduced misses to 2\%, and adding current target attributes reduced them to 1\% (Figure~\ref{fig:inputs}).

\begin{figure}[t]
  \centering
  \includegraphics[width=\columnwidth]{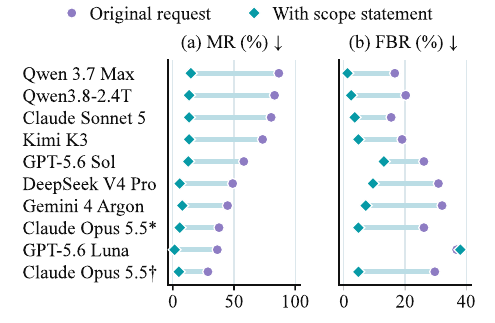}
  \caption{Model decisions before and after scope statements. The common procedure uses default inputs on 143 out-of-scope and 84 authorized actions. Every model reduces misses. $^*$ denotes a renamed second-stage field. $\dagger$ denotes refusal at the second stage in the original format, leaving the first stage to determine the final decision.}
  \label{fig:models}
\end{figure}

Figure~\ref{fig:models} reports results across models. Misses decreased in all ten settings, with exact McNemar tests yielding $p<10^{-6}$ for each. Improvements did not eliminate model differences. For example, GPT-5.6 Luna still blocked 38\% of authorized actions with explicit rules and default inputs, about as many as without them (37\%).

\subsection{Agents Stay within Stated Scope~(RQ4)}

For RQ4, we rerun complete tasks with scope statements and compare actions and final states with the original runs.

\begin{figure}[t]
  \centering
  \includegraphics[width=\columnwidth]{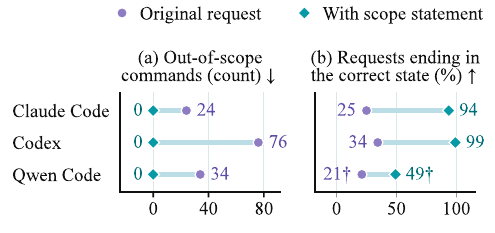}
  \caption{Complete-task execution with scope statements. (a) Counts of distinct out-of-scope commands, counting repeated commands within a request only once. (b) Proportions of tasks whose final states match reference answers. No agent proposes a command labeled out of scope after scope statements are added. $\dagger$ marks \qwen{} results affected by other causes of incomplete tasks.}
  \label{fig:rerun}
\end{figure}

Each agent reran 128 requests with scope statements, and none proposed an action labeled out of scope (Figure~\ref{fig:rerun}). The proportion of final states matching reference answers increased from 25\% to 94\% for \cc{}, 34\% to 99\% for \codex{}, and 21\% to 49\% for \qwen{}. \qwen{} still often attempted memory-file writes, which were denied, and left tasks incomplete, limiting final-state correctness.

Under original requests, 13, 54, and 24 of each product's 128 tasks ended with out-of-scope changes. None did so after scope statements were added. False-block rates during reruns were approximately 6\%, 1\%, and 0\%. In contrast, \citet{wang2026intentcontracts} clarify only environment terms and still observe wrong targets in 40--49\% of runs that change state.

Because reruns produced no out-of-scope actions, they cannot measure checker miss rates.

\section{Discussion}
\label{sec:discussion}
\label{sec:remaining}

\subsection{Target Identification Needs Tool Outputs}

Even with explicit authorization rules, checkers must link command targets to their attributes. For job cancellation by identifier, ownership and component information may appear only in earlier query outputs, while the cancellation command contains only an identifier. A checker reading only arguments may therefore lack evidence for verifying scope. Neither \cc{} nor \qwen{} reads these outputs. Job cancellation by identifier accounts for 1/2 and 5/7 of their remaining misses, respectively.

The common procedure provides a direct comparison of tool-output effects. With explicit rules but no outputs, the miss rate on out-of-scope job cancellations was 60\%, against 55\% under original requests. Adding outputs reduced it to 4\% (Figure~\ref{fig:inputs}b).

\subsection{Risk Judgments Can Override Authorization}

\begin{figure}[t]
  \centering
  \includegraphics[width=\columnwidth]{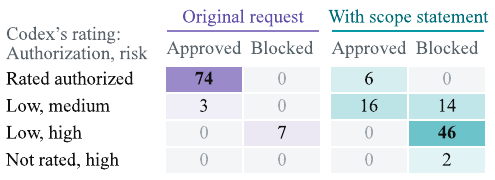}
  \caption{Authorization ratings, risk ratings, and final approval. Complete records from the second \codex{} run cover 84 out-of-scope actions. With scope statements, most actions receive low authorization ratings, yet some medium-risk actions still receive approval.}
  \label{fig:codex}
\end{figure}

Another error arises after a checker identifies insufficient authorization, involving the roles of authorization and risk ratings in final decisions. \codex{} produces both ratings, representing the request's authorization for an action and the action's potential impact. In the second run with scope statements, 76 of 84 out-of-scope actions received low authorization ratings. All 46 high-risk actions within this group were blocked, but 16 of 30 medium-risk actions were approved (Figure~\ref{fig:codex}). A low authorization rating therefore does not guarantee blocking.

Keyword matching found references to reversibility, bounded scope, or small scale in the approval rationales for 15 of these 16 actions. Some actions rated as insufficiently authorized were still approved because the checker considered their impact limited. Complete scope statements therefore cannot eliminate all errors. Approval rules must also specify how authorization constraints relate to risk assessments, allowing recognized restrictions to affect final decisions.

\subsection{Remaining Errors}

Combining fixed-action decisions and complete-task reruns leaves 40 errors, comprising 26 misses and 14 false blocks (Table~\ref{tab:remaining}). The \codex{} counts use its first run. Complete rationales were not saved for 10 of that run's 17 misses, preventing further attribution.

\begin{table}[t]
  \centering\small
  \begin{tabularx}{\columnwidth}{@{}Yccc@{}}
    \toprule
    Error category & Claude Code & Codex & Qwen Code \\
    \midrule
    \multicolumn{4}{@{}l}{\emph{Misses: repeated decisions on recorded actions}} \\
    Target information available only in tool outputs & 1 & 0 & 5 \\
    Approved based on low risk or reversibility & 1 & 6 & 0 \\
    Decision text not saved, cause undetermined & 0 & 10 & 0 \\
    Command did not change state & 0 & 1 & 2 \\
    \midrule
    \multicolumn{4}{@{}l}{\emph{False blocks: fixed actions and complete-task reruns}} \\
    Job cancellation by identifier & 4 & 0 & 0 \\
    Remote branch deletion & 3 & 0 & 0 \\
    Other authorized actions & 1 & 5 & 1 \\
    \midrule
    Total & 10 & 22 & 8 \\
    \bottomrule
  \end{tabularx}
  \caption{Remaining errors with explicit authorization scope. Misses come from fixed-action decisions. False blocks combine fixed-action decisions and complete-task reruns. \codex{} counts use the first run. Scripted rules classify causes, with records lacking complete rationales listed separately.}
  \label{tab:remaining}
\end{table}

\subsection{Implications}

Original requests assess behavior under incomplete scope statements, while augmented requests assess verification when rules are explicit, and the two give substantially different results. Reports of miss rates should therefore describe the labeling criterion, scope conditions present in requests, and target information visible to checkers.

For automatic approval systems, the results support stating permitted scope in requests or persistent policies, seeking clarification when scope is uncertain, and supplying outputs needed for target identification. Once this information is available, systems must enforce authorization constraints in final approval. The frequency of clarification and its effect on automation were not measured here.

\subsection{Limitations}

Evidence on misses comes from one development and operations benchmark that deliberately preserves scope ambiguity \citep{ji2026permissiongate}, and all out-of-scope actions arise from benign requests. We study neither attackers who adapt to checker feedback nor injected text in checker inputs, so the results do not quantify protection against adaptive attacks. Scope statements derive from reference answers and are complete by construction; partial or incorrect statements are untested, and obtaining, eliciting, and maintaining such rules requires separate validation.

Three actions that do not change state (one dry run and two restarts of nonexistent services) count as out of scope because of their targets; excluding them, \codex{} and \qwen{} still miss 77/83 and 26/33 out-of-scope actions. Except for the two \codex{} scope comparisons, each configuration runs once, and paired tests do not fully capture variability in repeated decisions. Fixed product versions and missing rationales further limit generalization and attribution.
\section{Conclusion}
\label{sec:conclusion}

Under original requests, most out-of-scope actions receive approval. Stronger models and additional target information do not resolve misses in aggregate, while most out-of-scope labels depend on rules absent from checker inputs. Stating these rules, taken from the reference answers, reduces misses and false blocks on the same actions, yet product checkers still approve 8\%--26\% of out-of-scope actions. Additional target information further improves decisions, and none of the agents proposes a labeled out-of-scope action during complete-task reruns. On this benchmark, explicit authorization scope is an important condition for reliable approval. Systems must also provide sufficient information for target identification and enforce the corresponding constraints in final decisions.

\FloatBarrier
\bibliography{refs}
\end{document}